\documentclass[final]{cvpr}

\usepackage{times}
\usepackage{epsfig}
\usepackage{graphicx}
\usepackage{amsmath}
\usepackage{amssymb}
\usepackage{caption}
\usepackage{subcaption}
\usepackage{siunitx}

\usepackage[pagebackref=true,breaklinks=true,colorlinks,bookmarks=false]{hyperref}

\def\cvprPaperID{1303} % *** Enter the CVPR Paper ID here
\def\confYear{CVPR 2021}

\begin{document}
%Commands
\newcommand{\norm}[1]{\left\lVert#1\right\rVert}
%\newcolumntype{C}{>{\centering\arraybackslash}p{3em}}
\renewcommand{\arraystretch}{1.3}
\setlength{\belowcaptionskip}{1pt}
\frenchspacing

%%%%%%%%% TITLE
\title{E3D: Event-based 3D Shape Reconstruction}

\author{
Alexis Baudron
\qquad
Zihao W. Wang
\qquad
Oliver Cossairt
\qquad
Aggelos K. Katsaggelos \\
Northwestern University \\
{\tt\small alexis.baudron@u.northwestern.edu}
}

\maketitle

%%%%%%%%% ABSTRACT
\begin{abstract}
    3D shape reconstruction is a primary component of augmented/virtual reality. Despite being highly advanced, existing solutions based on RGB, RGB-D and Lidar sensors are power and data intensive, which introduces challenges for deployment in edge devices. We approach 3D reconstruction with an event camera, a sensor with significantly lower power, latency and data expense while enabling high dynamic range. While previous event-based 3D reconstruction methods are primarily based on stereo vision, we cast the problem as multi-view shape from silhouette using a monocular event camera. The output from a moving event camera is a sparse point set of space-time gradients, largely sketching scene/object edges and contours. We first introduce an event-to-silhouette (E2S) neural network module to transform a stack of event frames to the corresponding silhouettes, with additional neural branches for camera pose regression. Second, we introduce E3D, which employs a 3D differentiable renderer (PyTorch3D) to enforce cross-view 3D mesh consistency and fine-tune the E2S and pose network. Lastly, we introduce a 3D-to-events simulation pipeline and apply it to publicly available object datasets and generate synthetic event/silhouette training pairs for supervised learning. 
\end{abstract}

%%%%%%%%% BODY TEXT
\section{Introduction}

Recent advances in 3D sensors have, more than ever, increased our access to 3D information, empowering high quality 3D reconstruction, scene/model recreation and human-machine interaction. One approach to 3D sensing is via active illumination. Such technologies include structured light, photometric stereo, time-of-flight and LiDAR, which have appeared in commodity devices such as Microsoft Kinect and Apple iPhone/iPad Pro, etc. Nonetheless, their high power consumption and inflexibility to fast motion have hindered broader applications in hand-held and head-mounted devices. Passive 3D does not incorporate additional illumination, but only explores photographs \cite{vslam, sfmrevisited}. The burden for passive photographic 3D, which typically involves multi-view registration and merging, relies heavily on the data quality, quantity, and computation, all of which pose significant challenges for augmented and mixed reality devices.

\begin{figure}
\centering
\begin{subfigure}{.16\textwidth}
  \centering
  \includegraphics[width=\textwidth]{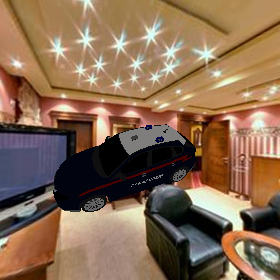}
  \caption{}
\end{subfigure}%
\begin{subfigure}{.16\textwidth}
  \centering
  \includegraphics[width=\textwidth]{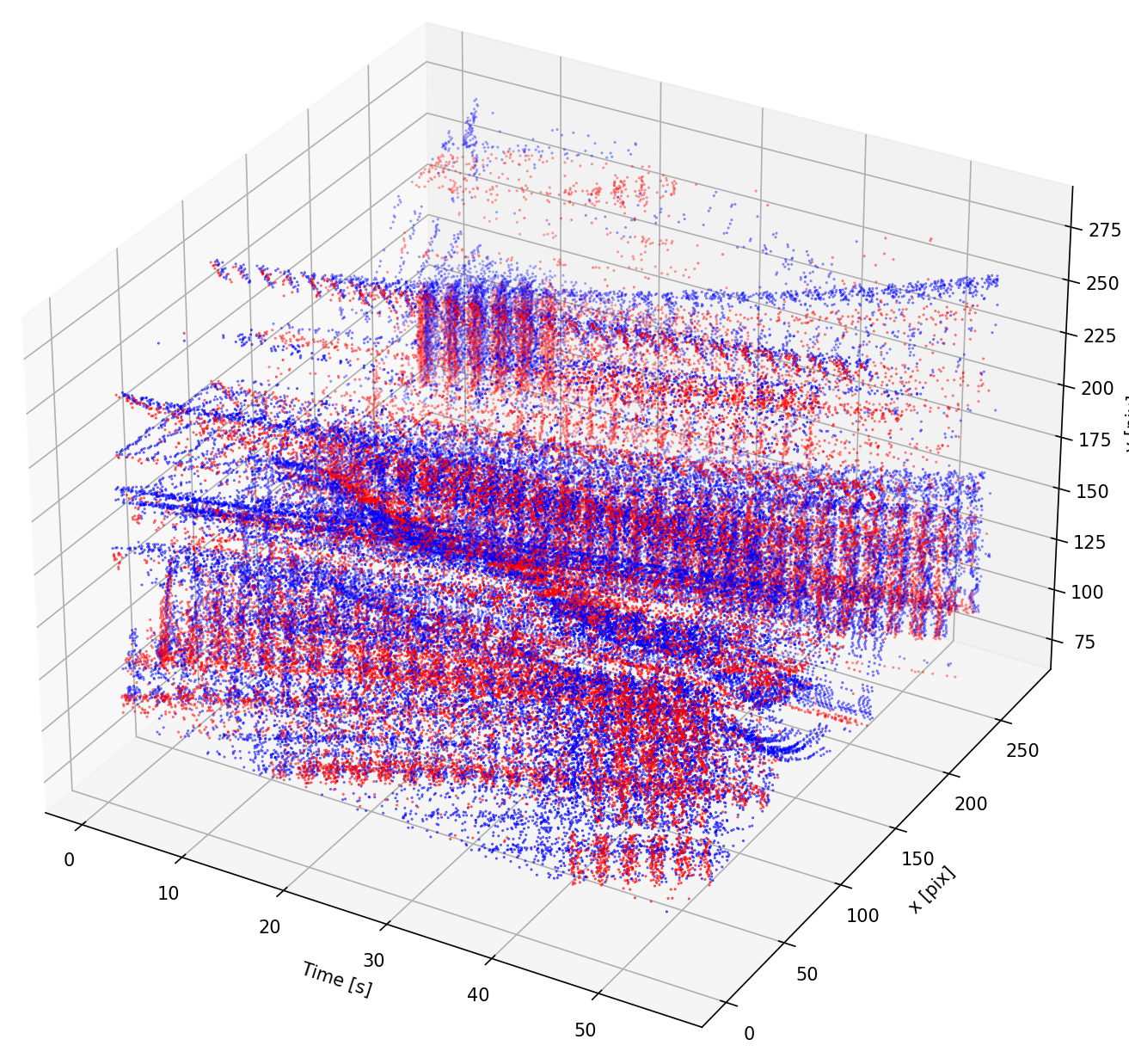}
  \caption{}
\end{subfigure}%
\begin{subfigure}{.15\textwidth}
  \centering
  \includegraphics[width=\textwidth]{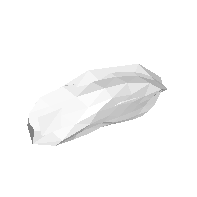}
  \caption{}
\end{subfigure}
\caption{a) Rendered image of a car overlayed on a textured background. b) Event volume caused by the viewer orbiting around the car (blue: positive event, red: negative event) c) Image of our reconstructed polygon mesh.}
\label{teaser}
\end{figure}

Event cameras are potential candidates for cost-effective 3D sensing, owing to their novel passive pixel design. Unlike mainstream active pixel sensors that produce image sequences by a clocked repetition of shutter control, event pixels perform state comparison and trigger event output only when strong temporal variation occurs. This, in conjunction with asynchronous pixel firing, has enabled low latency visual sensing with exceptionally low power and bandwidth constraints. Event cameras can be regarded as high speed sensors primarily measuring scene edges, following the close relation between temporal and spatial gradients. In the context of 3D information retrieval, methods have been introduced to convert event streams, which consist of spatio-temporal pointclouds, to sparse, semi-dense pointclouds \cite{kim2016, Rebecq2017EMVSEM}, or full-frame depth maps \cite{zhu2018unsupervised, ye2019unsupervised} through deep learning approaches.

In this work we focus on event-only 3D mesh reconstruction for rigid objects. Unlike previous stereo and feature-based methods, we approach dense 3D shape by establishing silhouette priors for multi-view shape reconstruction, see Fig.~\ref{teaser}. Silhouettes have been studied extensively as a means for recovering shape, from traditional Computer Vision techniques such as Shape-from-Silhouette \cite{Baumgart1974GeometricMF, visualhullLaurentini}, to modern deep learning approaches \cite{silnet2017, mvc2018}
and inverse graphics \cite{loper2014opendr, kato2017neural, liu2019soft}. 

We compare our results with video-based RGB approaches on the task of single-category shape reconstruction by reporting our reconstruction accuracy from events and contrasting our robustness to high-speed changes. We anticipate this work to have many potential applications in high-speed motion tracking, AR/VR and recovering 3D priors for downstream algorithms.

In summary, our main contributions are: 
\begin{itemize}
\item A novel algorithm for recovering a dense 3D mesh directly from events without the need for 3D supervision.
\item A novel supervision through mesh reconstruction loss applied to the task of joint silhouette and absolute pose prediction from event frames.
\item A synthetic dataset generation approach for object-specific learning tasks from events.
\item Qualitative and quantitative evaluations of our approach on single category ShapeNet objects.
\end{itemize}

\section{Related Work}

\subsection{Silhouette supervised shape reconstruction}
Shape-from-Silhouette (SfS) uses silhouettes of an object from different viewpoints to build the Visual Hull, the greatest intersection of the recovered silhouettes \cite{visualhullLaurentini}. This method was first introduced in 1974 by Baumgart in his PhD thesis \cite{Baumgart1974GeometricMF} and later extended \cite{projective_hulls, frontiers, sfs_acrosstime}. SfS is limited by the number of silhouettes used for reconstruction, outputting a very coarse approximation of the shape when there are few measurements. More recently, learning based approaches have shown to be successful in multi-view object reconstruction, using silhouettes as priors for recovering 3D shape \cite{silnet2017} or as supervision for learning shape \cite{tulsiani2017mvray, mvc2018}. Silhouettes are also used as reference measurements for multi-view shape-fitting of colourless meshes from differentiable renderering \cite{loper2014opendr, kato2017neural, liu2019soft, Pytorch3D}.
Similarly to SilNet \cite{silnet2017} we use silhouettes as priors for recovering shape from events while leveraging the multi-view optimization approach of inverse graphics. This allows us to incorporate any number of views into our reconstruction.

\subsection{Event-based multi-view 3D sensing}
Image-based reconstruction approaches such as simultaneous localization and mapping (SLAM) \cite{lsd-slam, orb-slam} and structure-from-motion (SfM) \cite{vslam, sfmrevisited} rely on feature-based correspondences across frames to recover camera pose and depth information. Events often pick up dominant features in the scene such as edges, making them ideally suited for feature-based methods. Kim et al. \cite{kim2016} propose three probabilistic filters capable of building a local map for camera localization. Rebecq et al. \cite{Rebecq2017EMVSEM} build semi-dense 3D depth maps from the event stream using feature-based tracking. Event cameras can also be coupled with RGB cameras to leverage the advantages of both sensors simultaneously. Vidal et al. \cite{vidal2018ultimate} create a SLAM pipeline that outperforms existing event-based methods by incorporating intensity frames, events and IMU data.
More recently neural networks have been trained to recover dense depth maps from events \cite{zhu2018unsupervised, ye2019unsupervised}. In \cite{zhu2018unsupervised}, Zhu et al. propose a novel discretized volumetric representation of events and use it to train an unsupervised convolutional network that predicts depth maps and camera motion from a stereo pair of event volumes. EventCap \cite{xu2019eventcap} proposes to use event cameras for fast human motion capture, but use intensity images to do human shape reconstruction. We introduce a method to infer dense 3D shape from silhouettes that are estimated directly from event data. In this paper, we focus on how to use object contour as a cue to estimate 3D shape, but the methods introduced can also be combined with other depth cues (e.g., texture and shading from intensity images) to improve 3D reconstruction results.

\section{Silhouettes as a proxy for recovering shape from events}
\begin{figure}[h!]
\includegraphics[width=.225\textwidth]{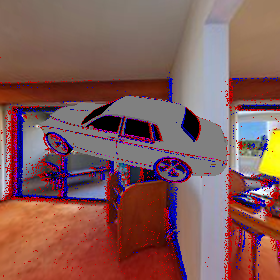}
\includegraphics[width=.225\textwidth]{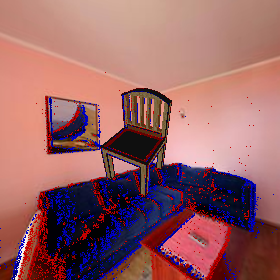}
\caption{Overlay of our event frames on top of synthetic images of objects with random textured backgrounds.} \label{contour generators}
\end{figure}
\begin{figure*}[t!]
  \centering
  \includegraphics[width=\textwidth, height=4cm]{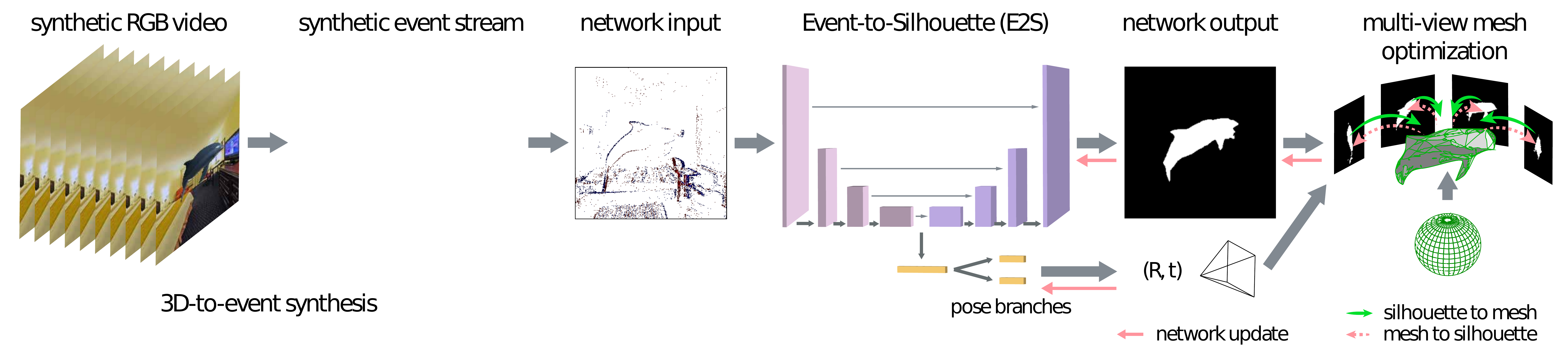}
  \caption{E3D architecture with mesh supervision. Our network learns to predict pose and object silhouettes from single event frames. The Event-to-Silhouette (E2S) and pose branches share the same encoder (in pink). The predicted silhouettes and poses are used to supervise our multi-view mesh optimization. The event frame is shown with a white background for increased visibility.}
  \label{E3D Pipeline}
\end{figure*}

Under constant viewer motion, events are triggered by the intensity contrast between an object's texture and the scene background (see Fig.~\ref{contour generators}). Amongst other information about the shape and texture of the scene, the event stream naturally contains view-dependent \textit{contour generators} \cite{koenderinkcontours} of objects without requiring any pre-processing. From these contours we can recover information about the silhouette, the dominant feature of an object on the image plane \cite{cipolla1992surface}. 

Our 3D reconstruction approach consists of first recovering silhouettes from the event stream through a CNN-based learning architecture. Our network learns to distinguish events that belong inside the contour generator and those that are part of the scene. By focusing on object silhouettes, instead of image intensities, we minimize the complexity of our network while also satisfying the constraint of 2D supervision only.

\section{E3D Approach}
\textbf{Overview} We consider the scenario where an event camera scans a static object in a random scene by rotating around it. We aim to reconstruct a 3D object mesh from a sequence of $N_E$ events $\{x_i, y_i, t_i, p_i\}, i \in N$, binned into a set of $N_B$ three-channel discretized event frames $E_{B}(x, y)$, described in \ref{Event Representation}. Our multi-view reconstruction approach, described in \ref{Mesh Model}, consists in optimizing the mesh $\hat{M}$ over learned silhouette priors $\hat{S}_{b}(x,y)$ and camera poses $\hat{\textbf{p}}_b = (\hat{t}_b, \hat{q}_b)$ predicted from each event frame $E_{b}(x, y)$ by the networks $f_s(\Theta_s)$ and $f_p(\Theta_p)$ respectively. We formalize the problem by the following equation: $\hat{M}$ = $G_m(f_s(E_{B}(x, y)), f_p(E_{B}(x,y))$, we define $G_m$ as the mesh optimization function. Our pipeline is described in Fig.~\ref{E3D Pipeline}.

\textit{Training Approach} As demonstrated in Fig.~\ref{E3D Pipeline}, we train our pipeline entirely in simulation by employing a 3D-to-event synthesis approach, described in greater detail in \ref{Synthetic Dataset}. Note, our approach uses 2D silhouette supervision only.

\textit{Geometric Supervision} During training we take a batch of predicted silhouettes $\hat{S}_{k}(x,y)$ and run them through our mesh optimization model $G_m$. From this we obtain an optimized mesh $M$, which we then project into 2D using a silhouette renderer $\pi$ for a given set of poses ${p_k}$:  $\hat{SM}_k(x,y)$ = $\pi(M | p_k)$ and take these silhouettes to supervise our Event-to-Silhouette network along with the predicted silhouettes $\hat{S}_{k}(x,y)$. Fig.~\ref{E3D Pipeline} showcases this process in action.

\subsection{Silhouette supervised object reconstruction}

\label{Mesh Model}
Given a set of silhouettes ${S_b}$ of an object and the respective camera extrinsics ${\textbf{p}_b}$ for each view, we aim to at least recover the visual hull, the greatest intersection of the visual cones, which in our case are the silhouettes predicted from the event stream. We formulate our approach as an inverse graphics energy minimization problem, given a set of reference views of an object and a differentiable silhouette renderer $\pi$, we offset the mesh $M$ so that it minimizes the loss function $\mathcal{L}(M)$. 

We start by initializing a template sphere ~\cite{kato2017neural}, composed of vertices ${v_i}$ and triangular faces ${f_i}$. At each forward step we deform the template mesh, render silhouettes for each known camera position on the trajectory and compare our 3D prediction with a small subset of the reference views. This approach allows us to maximize the cross-view geometric consistency without suffering from the drawbacks of the increased sampling rate of our object over an entire video. We deform the mesh while trying to minimize the following reconstruction losses:
\begin{itemize} \label{reconstruction losses}
    \item \textit{Silhouette Re-projection loss} or negative Intersection over Union (IoU) tries to maximize the overlap between the groundtruth image and the projected template mesh. The negative IoU is defined as in ~\cite{kato2017neural}:
    \begin{equation} \label{iou}
        \mathcal{L}_{IoU} = 1 - \frac{||S_{gt} \otimes S_{mesh}||_1}{||S_{gt} \oplus S_{mesh} - S_{gt} \otimes S_{mesh}||_1}
    \end{equation}
    \item \textit{Laplacian Loss} acts as a regularization term for the vertices by minimizing the distance between each vertex $v_i = (x_i, y_i, z_i)$ and its neighbors centers of mass. The Laplacian loss is defined in ~\cite{wang2018pixel2mesh}:
    \begin{equation}
    \begin{aligned}
    \delta_i &= v_i - \frac{1}{||N(i)||} \sum_{j\in N(i)}v_j \\
    \mathcal{L}_{lap} &= ||\delta_i||_2^2
    \end{aligned}
    \end{equation}
    \item \textit{Flattening Loss} further smooths out the mesh surface and prevents self-intersections by minimizing surface normal difference between a face and its neighbors. We define $\theta_i$ as the angle between face $f_i$ and the surrounding faces that share the same edge $e_i$. The flattening loss is defined as in ~\cite{kato2017neural}:
    \begin{equation}
    \mathcal{L}_{smooth} = \sum_{\theta_i\in e_i} (cos\theta_i + 1)^2
    \end{equation}
\end{itemize}
The objective function is a weighted sum of these three losses: $\mathcal{L}(M)$ = $\lambda_{IoU}*\mathcal{L}_{IoU}$ + $\lambda_{lap}*\mathcal{L}_{lap}$ + $\lambda_{smooth}*\mathcal{L}_{smooth}$. 

\subsection{Event Representation} \label{Event Representation}
Event cameras respond to changes in a scene's brightness, with each pixel on the event camera responding asynchronously and independently. This processing approach is inspired by human vision by modelling visual pathways in the human brain. Unlike a shutter-based RGB camera, which captures all incoming light at pre-determined time intervals, an event camera outputs a stream of events encoding the thresholded log intensity change at each pixel. Events are fired when their intensity is above the contrast threshold CT:
\begin{equation}
    \pm CT = logI(x, y, t) - logI(x, y, t - \Delta t)
\end{equation}
Each event $(e_k)$ encodes the pixel location $(x_k, y_k)$, the timestamp $(t_k)$ in $\mu$s at which it was fired and the intensity polarity $p_k \in \{ -1, 1\}$
\begin{equation}
    e_k = \{x_k, y_k, t_k, p_k\}
\end{equation} 

Our chosen event representation is inspired by ~\cite{EvSteeringPrediction2018} and meant to create event frames that are suitable for 2D learning architectures. Our count-based binning strategy ensures every frame has the same event density. From a sequence of N events $\{x_i, y_i, t_i, p_i\}, i \in N$, we create a set of $B = \frac{N}{count}$ where the count is a variable we control to create $B$ equal frames. For the creation of our synthetic dataset (see \ref{Synthetic Dataset}) we set the count so that we synthesize an equal amount of event frames as there are input images. We accumulate the events in a pixel-wise manner over the integration time interval $W$ of each bin to create 2D histograms $H$ for both positive and negative events at each pixel:
\begin{equation}
    H(x, y, p) = \sum_{i = 0, t_i \in W}^{N_{b}}\delta(x_i, x)\delta(y_i, y)\delta(p_i, p) 
\end{equation}
Each event frame contains the stacked histograms for both positive and negative polarities along with a third channel to create a three-dimensional dense encoding of the event stream. 

\subsection{Learning object silhouettes and absolute pose from events}
\subsubsection{Silhouette Prediction - E2S} \label{E2S}
The task of recovering object masks from event frames is the same as that of single-class semantic segmentation where we classify each pixel in the output frame as being in or out of the object. We use a CNN-based architecture to approach this problem given its success with RGB images and its resilience to sparse event data \cite{EvSegNet2018}. Given a deblurred event frame, we train our network to recover the object mask through ground truth supervision to which we apply a common binary cross entropy loss :
\begin{equation}
\mathcal{L}_{bce} = -(S * log(\hat{s})+(1-S)log(1-\hat{s})) 
\end{equation}
$S$ is the groundtruth silhouette of the object and $\hat{s}$ is an image composed of the pixel-wise probabilities of belonging to the silhouette.

\subsubsection{Joint learning of mask and pose - E2S + pose} \label{E2S + pose}
As shown in Fig.~\ref{E3D Pipeline} we extend our silhouette prediction network with a branch to regress camera pose $\textbf{p} = (t, q)$, $t \in \mathcal{R}^3$ and the quaternion parameterization of rotation $q \in \mathcal{R}^4$. As in \cite{kendall2016posenet}, we learn both the translation and orientation of the pose jointly. The network learns poses that are geometrically enforced using losses applied to per event-frame absolute pose $\mathcal{L}_{abs}$ and relative pose between event frame pairs $\mathcal{L}_{rel}$. The absolute loss between the groundtruth pose $\textbf{p} = (t, q)$ and the predicted pose $\hat{\textbf{p}} = (\hat{t}, \hat{q})$ for an event frame $E_i$ is defined as in ~\cite{geometricPoseNet}:
\begin{equation}
\begin{split}
    \mathcal{L}_{abs} &= h(p_i, \hat{p_i}) \\
    &= \norm{t_i - \hat{t_i}}_1e^{-\beta} + \beta * \norm{q_i - \hat{q_i}}_1e^{-\gamma} + \gamma
\end{split}
\end{equation}
We use the definition of $\mathcal{L}_{rel}$ described in \cite{MapNet2018} as the absolute loss applied to the predicted relative camera pose between $\hat{p_i}$ and $\hat{p_k}$, $\hat{r}_{ik} = (\hat{t}_i - \hat{t}_k, \hat{w}_i - \hat{w}_k)$ and the groundtruth relative camera pose $r_{ik}$:
\begin{align*}
    \mathcal{L}_{rel} = h(r_{ik}, \hat{r}_{ik})
\end{align*}
The optimal weighting of $\beta$ and $\gamma$ is learned during training.

\subsubsection{Supervision through shape reconstruction} \label{geometric segnet}
Recovering silhouettes with a naive binary cross entropy loss, as described in \ref{E2S}, does not incorporate temporal or spatial consistency. Different image-based video object segmentation approaches have proposed incorporating this information through temporal aggregation \cite{huang2018efficientseg}, foreground appearance propagation \cite{osvos2018} and foreground-background matching \cite{yang2020collaborativeseg}. We leverage the spatial consistency of the object by applying cross-view consistency through our mesh optimization function $G_m$.

During training, a batch of predicted silhouettes is used to create a mesh $M$ as described in \ref{Mesh Model}. The resulting mesh is then projected into silhouettes using the silhouette renderer $\pi$. We then use these silhouettes to update the weights of our network $f_s(\Theta_s)$ by comparing the projected silhouettes to the groundtruth using an Intersection-over-Union loss:
\begin{align*}
    \mathcal{L}_{shape} &= \sum_{b=0}^{B}\mathcal{L}_{IoU}(S_b, \pi(\hat{M} | \textbf{p}_b))
\end{align*}
The mesh optimization is a computationally expensive operation so we only use it as a fine-tuning step at the end of our training.

Our learning objective for the event-to-silhouette and pose network is: 
\begin{equation}
    \mathcal{L}_{E2S} = \mathcal{L}_{bce} + \mathcal{L}_{abs} + \mathcal{L}_{rel} + \mathcal{L}_{shape} * \lambda_{shape}
\end{equation}

\section{Synthetic Dataset Generation} \label{Synthetic Dataset}
Training neural networks requires large labelled datasets which are time intensive to collect. One option would be to rely on existing datasets, however, to the best of our knowledge there are no datasets of event-based category-specific object sampling that contain both groundtruth 2D object masks and camera poses. To avoid the cumbersome task of collecting and labelling our own event dataset, we propose training our networks in simulation by creating synthetic event data. 

\begin{figure}[h!]
  \includegraphics[width=.5\textwidth]{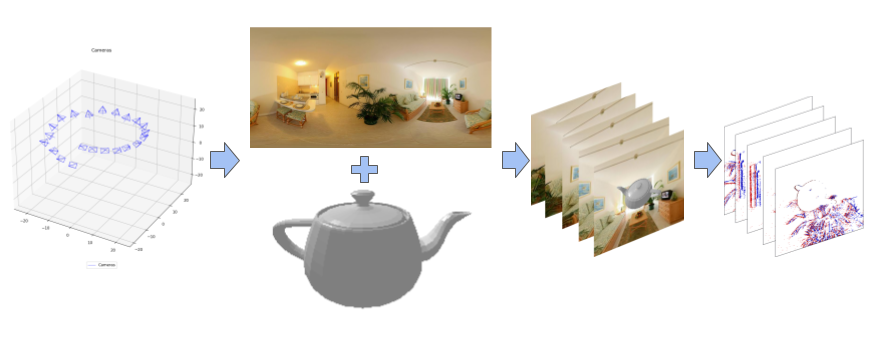}
  \caption{3D-to-event synthesis from trajectory generation to synthetic event frames}
  \label{synth-data}
\end{figure}

We generate realistic event data using the Event Simulator ESIM ~\cite{rpg_vid2e}, while obtaining the labels needed for training our architecture: groundtruth object silhouette and camera pose. To create the event data, we simulate a camera rotating around static textured object meshes and render an image at each viewpoint, giving us a video of RGB images. To ensure realism, we use textured backgrounds from SUN360 ~\cite{SUN360}, a database of 360 degree panoramas of indoor scenes which we then split into 360 perspective images through spherical projection. We use meshes from the ShapeNet Dataset ~\cite{ShapeNet} and build category specific event streams, from which we synthesize event frames as described in \ref{Event Representation}. Our synthetic event generation process is demonstrated in Fig.~\ref{synth-data}.

\textbf{Augmentation} We augment our dataset with unexpected camera motion and sensor noise to minimize the sim-to-real gap. A viewer rotating around an object while holding a camera would do a poor job at matching the smooth trajectory we could create in simulation. To address this discrepancy we introduce random starting points, long continuous oscillations along distance and elevation and micro-fluctuations along elevation and azimuth. Unlike \cite{li2018interiornet}, our trajectory augmentation is not learned, but approximates handheld motion in the controlled setting of orbiting around the object. We follow \cite{simtoreal} and add noise to the event stream in the form of zero mean gaussian noise $\mathcal{N}$(0,0.1) which simulates uncorrelated background noise. Additionally, we extend the implementation of ~\cite{rpg_vid2e} by simulating a varying contrast threshold by sampling it according to $\mathcal{N}$(C,$\sigma _c$) where $\sigma_c$ is a modifiable parameter as described in ~\cite{Rebecq18corlESIM}, we set $\sigma _c$ to be (0.03, 0.03) for positive and negative CTs. Additionally the mesh is randomly translated off-center before rendering.

\begin{figure*}[t]
\begin{tabular}{cccccc cc}
\includegraphics[width=.115\linewidth]{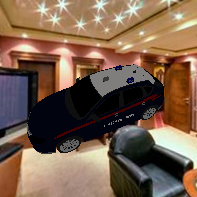} 
\includegraphics[width=.115\linewidth]{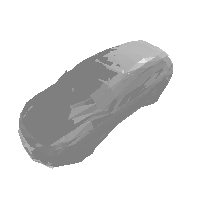} 
\includegraphics[width=.115\linewidth]{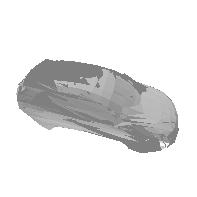} & 
\includegraphics[width=.115\linewidth]{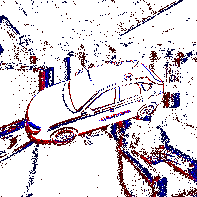}  
\includegraphics[width=.115\linewidth]{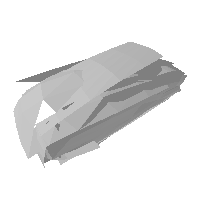}
\includegraphics[width=.115\linewidth]{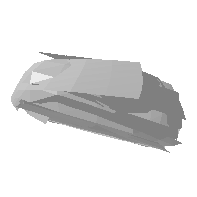} &
\includegraphics[width=.115\linewidth]{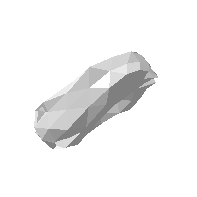}
\includegraphics[width=.115\linewidth]{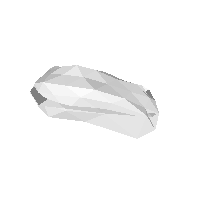}
\\
\includegraphics[width=.115\linewidth]{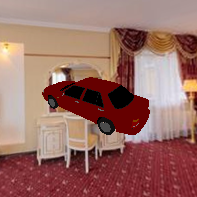}
\includegraphics[width=.115\linewidth]{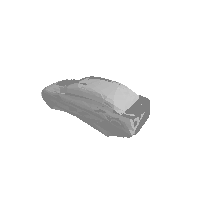}
\includegraphics[width=.115\linewidth]{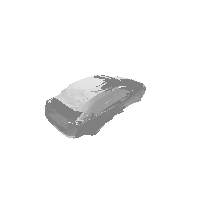}&
\includegraphics[width=.115\linewidth]{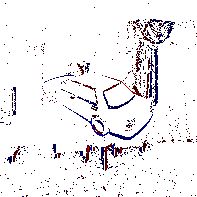}
\includegraphics[width=.115\linewidth]{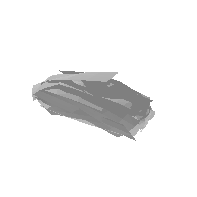}
\includegraphics[width=.115\linewidth]{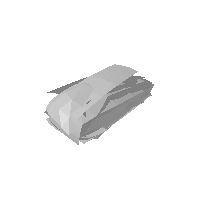} &
\includegraphics[width=.115\linewidth]{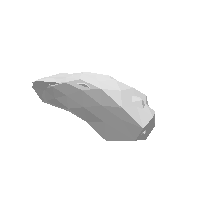}
\includegraphics[width=.115\linewidth]{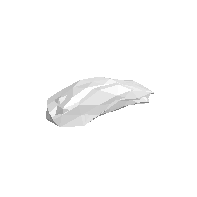}
\\ 
\includegraphics[width=.115\linewidth]{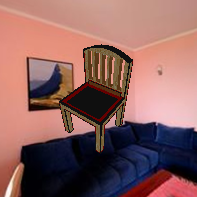} 
\includegraphics[width=.115\linewidth]{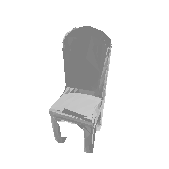}
\includegraphics[width=.115\linewidth]{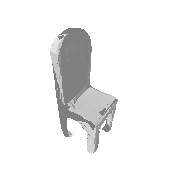} &
\includegraphics[width=.115\linewidth]{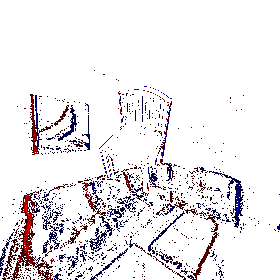}
\includegraphics[width=.115\linewidth]{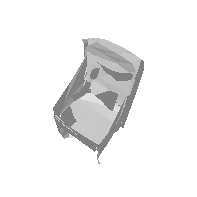} 
\includegraphics[width=.115\linewidth]{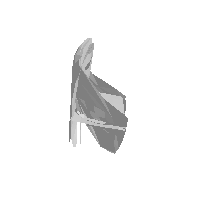} &
\includegraphics[width=.115\linewidth]{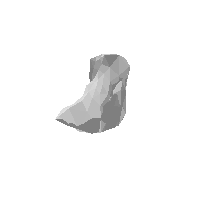}
\includegraphics[width=.115\linewidth]{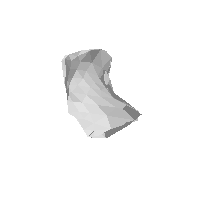}
\\ 
\includegraphics[width=.115\linewidth]{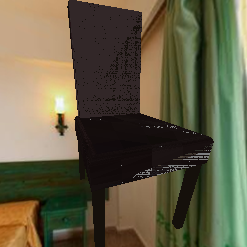} 
\includegraphics[width=.115\linewidth]{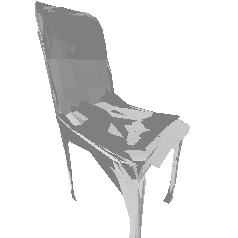}
\includegraphics[width=.115\linewidth]{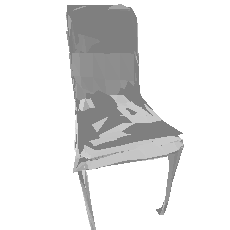} &
\includegraphics[width=.115\linewidth]{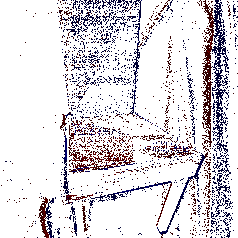}
\includegraphics[width=.115\linewidth]{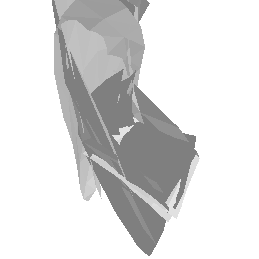} 
\includegraphics[width=.115\linewidth]{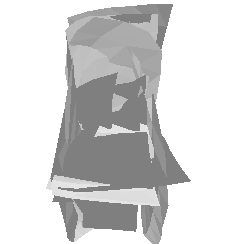} &
\includegraphics[width=.115\linewidth]{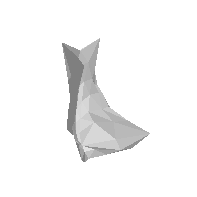}
\includegraphics[width=.115\linewidth]{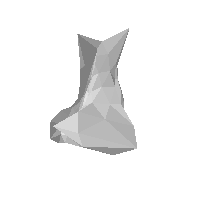}
\\ 
\end{tabular}
\caption{Single Category results on ShapeNet results, we show two views for each reconstruction. From left to right: RGB sequence, PMO ~\cite{videomesh2019} (2 views), Event Frame, PMO re-trained on events (2 views) and E3D (2 views).} \label{Shapenet Qual}
\end{figure*}

\begin{table*}[h]
\begin{center}
\resizebox{.8\textwidth}{!}{
\begin{tabular}{|c|c|c|c|c|c|c|c|c|c|}
\hline
 & \multicolumn{7}{c|}{\textbf{E3D (Ours)}} & \multicolumn{2}{c|}{\textbf{PMO}}  \\ \hline
 Category & Sil & \multicolumn{2}{c|}{GT Pose} & \multicolumn{2}{c|}{Trans Only} & \multicolumn{2}{c|}{Rot \& Trans} & original & events\\ \hline
  & \textit{\small IoU} & \textit{\small IoU} & \textit{\small Chamfer} & \textit{\small IoU} & \textit{\small Chamfer} & \textit{\small IoU} & \textit{\small Chamfer} & \textit{\small Chamfer} & \textit{\small Chamfer} \\ \hline\hline
 Car & .34 & .27 & 1.6 & .43 & 2.5 & .60 & 20.8 & .7 & 21.3\\ 
 Chair & .45 & .35 & 2.7 & .46 & 3.8 & .71 & 25.0 & 2.3 & 22.6\\ \hline
 Mean  & .39 & .31 & 2.1 & .44 & 3.1 & .65 & 22.9 & 1.5 & 21.95\\ \hline
\end{tabular}
}
\caption{Mesh quantitative results on ShapeNet dataset, we report our reconstruction results for both Mean IoU and Chamfer distance. We also report the negative mean IoU for our silhouette prediction. For the original PMO results, we scale by $10^{3}$ for readability. For negative IoU, lower is better.} \label{shapenet mesh}
\end{center}
\end{table*}

\section{Experiments}
\textbf{Evaluation Metrics} \label{evaluation metrics}
We test the different parts of our pipeline using test sets composed of images previously unseen by the network. For each event video in the test set we run our pipeline and measure the silhouette, camera pose and mesh accuracy. \par
For the segmentation accuracy of our E2S network we measure the overlap between the predicted object mask and the groundtruth mask $sil_{err}$ generated by our synthetic dataset using the IoU loss defined in eq.~\ref{iou}. We report our score as a mean IoU averaged over every prediction and groundtruth pair $(S, \hat{S})$. \par
For camera pose, we evaluate both the translation error and angular error independently by comparing to our groundtruth camera pose given by the inverse of the camera's world-to-view transform in Pytorch3D. We normalize the quaternion to unit length at test time. Given a groundtruth pose $p = (t, q)$ and a predicted pose $\hat{p} = (\hat{t}, \hat{q})$, we measure the translation error in meters as the euclidean distance: $t_{err} = \delta x = \norm{t - \hat{t}}_2$ and the rotation error in degrees as the minimum rotation angle $\alpha$ required to align the groundtruth and estimated rotations: $rot_{err} = \alpha = 2 \cos^{-1}{|q\hat{q}|}\frac{180}{\pi}$. \par
To measure the quality of our mesh reconstruction we report two values: mean IoU and 3D chamfer distance. We calculate the mean negative IoU of the silhouette projection of the predicted mesh silhouettes $\hat(s)_M$ at a given camera pose $p$ using a silhouette renderer $\pi$: $M_{iou} = IoU(\pi(\hat(M), p), S_{gt}$, allowing us to measure the mesh's quality at different viewpoints. We also report 3D chamfer distance between predicted mesh $\hat{M}$ and the groundtruth mesh $M$. Finally, we sample points on the surface of each mesh to create a pointcloud representation of the mesh and calculate the mean distance across all points. 

\textbf{Implementation Details} The Event-to-Silhouette (E2S) network is optimized with RMSProp, the learning rate is set to $10^{-4}$, weight decay of $10^{-8}$ and momentum $9 * 10^{-1}$. We set the threshold confidence for the segmentation predictions during training at $5 * 10^{-1}$. The pose branches are optimized using ADAM \cite{adam2017} with a learning rate of $10^{-4}$ and weight decay $5 * 10^{-4}$. The training examples are randomly sampled from each object set to ensure the network looks at one object at a time and we set the batch size to 6. During fine-tuning we set $\lambda_{shape} = 1.2$. The mesh reconstruction is optimized using ADAM with a learning rate of $5 * 10^{-3}$ and betas (0.9, 0.99). The silhouette renderer used to rasterize the mesh is initialized with a blur radius $\log(1 / 10^{-4} - 1) * 10^{-4}$ to be consistent with ~\cite{liu2019soft}. For mesh reconstruction we use the following weights $\lambda_{IoU} = 1.0$, $\lambda_{lap} = 1.0$ and $\lambda_{smooth} = 10^{-2}$

\subsection{ShapeNet Evaluation}

We use the ShapeNet dataset \cite{ShapeNet} to empirically validate our approach on single-object categories. We evaluate on two categories commonly used in other 3D reconstruction papers \cite{videomesh2019, mvc2018, gkioxari2019mesh}: cars and chairs. We use a subset of these categories composed of 433 and 859 objects respectively. For each object we follow the process described in \ref{Synthetic Dataset} and save the event frames and labels from 45 viewpoints with a resolution of 280x280. For ShapeNet, we train the E2S network and pose branch simultaneously until the loss converges. We create a random train/test split 80\%-20\% and during training we reserve 10\% of the train split for validation. At test time, we predict a silhouette and camera pose for each of the event frames in the test set. At each iteration of the mesh optimization, we randomly select two predicted silhouettes and use them to calculate our re-projection loss $\mathcal{L}_{IoU}$.

To the best of our knowledge, there are no papers that tackle object mesh reconstruction directly from events that would allow us to perform a direct comparison. We decide to compare our results on single-object mesh reconstruction with Photometric Mesh Optimization (PMO) ~\cite{videomesh2019}, which tackles RGB video mesh reconstruction. It relies on the MLP-based AtlasNet ~\cite{groueix2018atlasnet} as a shape generating decoder to make an initial guess at a selected view and improves it using cross-view photometric consistency. AtlasNet is pre-trained on ShapeNet using groundtruth pointcloud supervision. Although this method is partially 3D supervised while we only use 2D supervision, it is a close comparison due to its video-based optimization approach. We retrain PMO using our synthetic event dataset for 700 epochs for both the car and chair categories. We show qualitative results in Fig.~\ref{Shapenet Qual} and also compare with the original implementation of PMO. Our event-based method outperforms the retrained PMO. Although it is capable of capturing the overall shape, some of the vertices degenerate into unrealistic locations. This could be because of two factors: the initial guess made by AtlasNet is too noisy to recover from, or the regularization terms included by PMO do not enforce mesh consistency as strongly as our regularization terms $\mathcal{L}_{smooth}$ and $\mathcal{L}_{lap}$ (\ref{reconstruction losses}). In comparison to the image-based PMO, our method does poorly at recovering small details such as the legs of the chair. These small details are not present in our predicted silhouettes because of the sparsity of events being triggered on these contours. \par

We evaluate our results quantitatively on our synthetic event ShapeNet dataset and show our results for both mesh and pose. Table.~\ref{shapenet mesh} shows our reconstruction results compared with those of PMO. In addition to reporting the quality of our mesh reconstruction with groundtruth pose information, we report the results for: predicted translation \& orientation, and translation only. For translation only, we calculate the orientation of the camera by pointing it to the origin of our coordinate system. We report the chamfer distance results of both the original PMO trained using images and PMO re-trained on events. Our method outperforms PMO re-trained on events under both known camera pose and translation only settings, see Table.~\ref{shapenet mesh}. Note, PMO assumes known camera pose for their experiments. Table.~\ref{shapenet pose} shows our pose accuracy and error results on our dataset, these results are for single-view pose estimation. Interestingly, our mesh reconstruction quality improves when using the translation only approach, yet the orientation error increases. We attribute this to the off-center translation of the mesh and orientation prediction outliers.

\begin{table}[h]
\begin{center}
\resizebox{0.45\textwidth}{!}{
\begin{tabular}{c|c|c|c|c|c}
 Category & \multicolumn{3}{c|}{Rot $\&$ Trans} & \multicolumn{2}{c}{Trans only}\\ 
  & \multicolumn{2}{c|}{\textit{\small Median Error}} & \textit{\small Accuracy} & \multicolumn{2}{c}{\textit{\small Median Error}}\\
  & $t_{err}[m]$ & $q_{err}[\si{\degree}]$ & \small $<$1m,45$\si{\degree}$ & $t_{err}[m]$ & $q_{err}[\si{\degree}]$\\ \hline
 Car & .4 & 38.3 & 74.4$\%$ & .4 & 49.0  \\ 
 Chair & 0.41 & 42.9 & 38$\%$ & .41 & 46.8 \\ \hline
 Mean & 0.40 & 40.6 & 56$\%$ & .40 & 47.9 \\ 
\end{tabular}
}
\end{center}
\caption{Pose quantitative results on ShapeNet dataset, we report median angular error and translation error (see \ref{evaluation metrics})}. \label{shapenet pose}
\end{table}

\begin{figure}[h!]
  \centering
  \includegraphics[width=.14\textwidth]{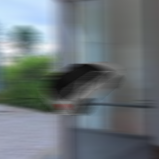}   \includegraphics[width=.14\textwidth]{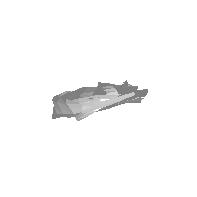}   \includegraphics[width=.14\textwidth]{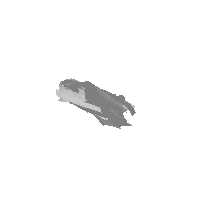}  
  \\  
  \includegraphics[width=.14\textwidth]{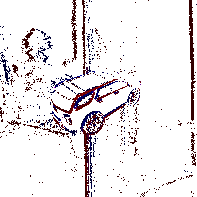}   \includegraphics[width=.14\textwidth]{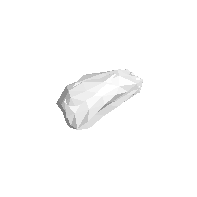}   \includegraphics[width=.14\textwidth]{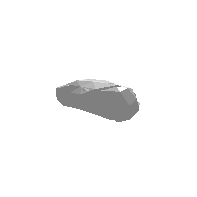} 
  \caption{Qualitative Motion Blur experiment. We show the meshes from two different views.} \label{motion blur}
\end{figure}

\subsection{Motion Blur Comparisons}
Motion blur can occur in shutter-based sensors when the scene changes during exposure. This can lead to serious degradation of the image, often impacting prediction results. The event camera has extremely low temporal resolution in the area of 1$\mu$, making it resilient to motion blur. Fig.~\ref{motion blur} shows qualitative results in comparison with PMO ~\cite{videomesh2019} running on frames that have been blurred using a directional filter to simulate the effect of motion blur. To simulate a high-speed capture environment for the event camera, we increase the elapsed time between sampled image frames, effectively speeding up our synthetic capture. The events preserve the visual content of the scene and do not suffer from motion blur.

\begin{table}[h!]
\begin{center}
\begin{tabular}{ c|c|c} 
Architecture & \multicolumn{2}{c}{Mean IoU Results}\\
& Segmentation & Mesh\\
\hline
GT & 0.0 & .10  \\ 
Fine-Tuned & .36 & .30 \\
Baseline & .37 & .32\\ 
 \hline
\end{tabular}
\end{center}
\caption{Quantitative results for our Loss function comparison. The network is trained on a scaled down object dataset of dolphins.} \label{loss comp}
\end{table}

\begin{figure}[h!]
  \centering
  \includegraphics[width=.14\textwidth]{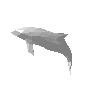}   \includegraphics[width=.14\textwidth]{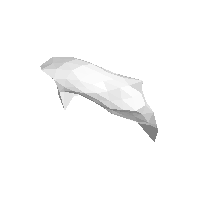}   \includegraphics[width=.14\textwidth]{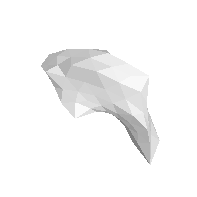}  
  \caption{Visual comparison of the effect of our loss on reconstruction quality. From left to right: GT, Fine-Tuned, Baseline. Although our baseline captures the shape of the dolphin (the IoU is still good) the shape of the dolphin is far more accurate in our fine-tuned reconstruction.} \label{loss-comp-visual}
\end{figure}

\subsection{Loss function Comparison}
In Table.~\ref{loss comp} we show quantitative results of the impact of our function introduced in \ref{geometric segnet}. Including this step from the start of training is a costly operation and would slow down our training time drastically. Instead, we fine-tune the network that has already been trained to recover the camera pose and silhouette of the object. This additional step refines silhouette and mesh predictions (Fig.~\ref{loss-comp-visual}) by enforcing the spatial consistency of the recovered object. We also include results of our mesh reconstruction using groundtruth silhouettes as a reference point for the quality of mesh reconstruction. Note, our mesh silhouette IoU improves in comparison to the input silhouettes because of our multi-view optimization approach.

\begin{figure}[h!]
  \centering
  \includegraphics[width=.19\textwidth]{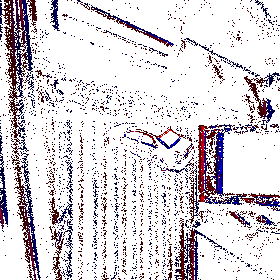}  \includegraphics[width=.19\textwidth]{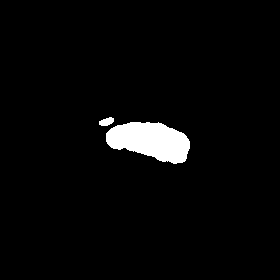} \\  
  \includegraphics[width=.19\textwidth]{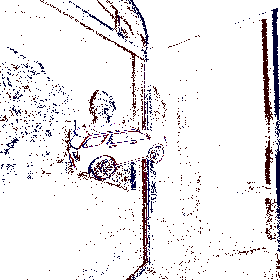}  \includegraphics[width=.19\textwidth]{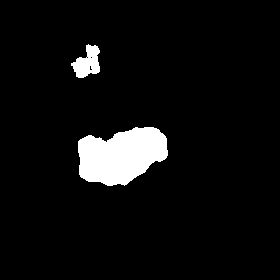}
  \caption{Silhouette predictions from our segmentation network E2S, noise appears in the predicted silhouette, introduced by the neural network during upsampling of the feature embeddings.}
  \label{failure}
\end{figure}

\section{Discussion and Limitations}
In this work we introduce E3D, a method to recover dense object meshes from sparse events. We first introduce an event-to-silhouette (E2S) network to recover object silhouettes from a stack of event frames. We extend E2S with additional branches for absolute camera pose regression from appearance-based cues. Finally, we use a differentiable renderer (Pytorch3D) to perform multi-view mesh optimization using our predictions.

Although our method outperforms a comparable image-based state-of-the-art method on recovering shape from events, it relies entirely on the quality of our silhouette predictions. These can often have noise in the form of chequered artifacts as shown in Fig.~\ref{failure}. The cross-view mesh optimization we adopt reduces our sensitivity to such artifacts. However, it can still have an impact on our overall reconstruction, see Table.~\ref{loss comp} for a comparison of our 3D reconstruction using predicted and groundtruth silhouettes. We intend to investigate additional temporal and spatial consistency constraints in follow-up work.

Our network learns to predict both translation and orientation from a single event frame. While our translation predictions allows to achieve good mesh reconstructions (Table.\ref{shapenet mesh}), our orientation predictions struggle to achieve good results. In future work we intend to address this issue through geometric losses such as re-projection error.

Despite these limitations, we are confident this approach will open future research into event-based shape reconstruction. Our method clearly shows high quality results which can open doors for using event cameras for dense 3D. 
To encourage follow-up work in object-centric 3D event sensing, we will make our code available.

{\small
\bibliographystyle{ieee_fullname}
\bibliography{egbib}
}

\end{document}